\documentclass[10pt,twocolumn,letterpaper]{article}

\usepackage{cvpr}              
\usepackage{graphicx}
\usepackage{amsmath}
\usepackage[symbol]{footmisc}
\usepackage{amssymb}
\usepackage{booktabs}
\DeclareMathOperator*{\argmax}{argmax}
\usepackage{algorithm}
\usepackage[noend]{algpseudocode}
\makeatletter
\def\BState{\State\hskip-\ALG@thistlm}
\makeatother

\newcommand{\algorithmicforeach}{\textbf{foreach}}

\algdef{SE}[FOREACH]{ForEach}{EndForEach}[1]{\algorithmicforeach\ #1\ \algorithmicdo}{\algorithmicend\ \algorithmicforeach}%
\algtext*{EndForEach}

\usepackage[pagebackref,breaklinks,colorlinks]{hyperref}

\usepackage[capitalize]{cleveref}
\crefname{section}{Sec.}{Secs.}
\Crefname{section}{Section}{Sections}
\Crefname{table}{Table}{Tables}
\crefname{table}{Tab.}{Tabs.}

\def\cvprPaperID{7769} 
\def\confName{CVPR}
\def\confYear{2022}

\newcommand*{\affaddr}[1]{#1} 
\newcommand*{\affmark}[1][*]{\textsuperscript{#1}}
\newcommand*{\email}[1]{\texttt{#1}}

\begin{document}

\title{Weakly-Supervised Online Action Segmentation in Multi-View \\ Instructional Videos}

\author{Reza Ghoddoosian\affmark[1,2]\thanks{Work done during Reza’s internship at Honda Research Institute, USA} , \  Isht Dwivedi\affmark[1], Nakul Agarwal\affmark[1], Chiho Choi\affmark[1], and Behzad Dariush\affmark[1]\\
\affaddr{\affmark[1]Honda Research Institute, USA}\\
\affaddr{\affmark[2]VLM Lab, University of Texas at Arlington}\\
\email{\small reza.ghoddoosian@mavs.uta.edu, \small \{idwivedi,nagarwal,cchoi,bdariush\}@honda-ri.com}
}



\maketitle
\begin{abstract}
 This paper addresses a new problem of weakly-supervised online action segmentation in instructional videos. We present a  framework to segment streaming videos online at test time using Dynamic Programming and show its advantages over greedy sliding window approach. We improve our framework by introducing the Online-Offline Discrepancy Loss (OODL) to encourage the segmentation results to have a higher temporal consistency. Furthermore, only during training, we exploit frame-wise correspondence between multiple views as supervision for training weakly-labeled instructional videos. In particular, we investigate three different multi-view inference techniques to generate more accurate frame-wise pseudo ground-truth with no additional annotation cost. We present results and ablation studies on two benchmark multi-view datasets, Breakfast and IKEA ASM. Experimental results show efficacy of the proposed methods both qualitatively and quantitatively in two domains of cooking and assembly.

\end{abstract}

\section{Introduction}
Action understanding in untrimmed instructional videos is important in many applications, where agents learn by observation of other agents performing complex tasks. Such videos are characterized by composition of a sequence of low-level atomic actions, e.g., \textit{crack eggs} and \textit{whisk eggs}, that form a high-level task, e.g., \textit{making eggs}. This contextual dependency between actions as well as other attributes in instructional videos have inspired new research~\cite{assessment,Cooking,COIN,chang2019procedure,HM_reza} that has advanced the field.

A fully-supervised training of these videos would require not only the labels for each action, but also their temporal assignment (start and end time) with ordering constraints. 
However, creating fully annotated clips with action assignments and labels on the temporal boundaries of individual actions is manually intensive and is therefore both time consuming and expensive.  This limits the scale and practicality at which fully-supervised video datasets can be created.  Furthermore, the subjective nature of labeling the start and end time of each action results in ambiguities and inconsistencies. In \textit{weakly-supervised action segmentation} these limitations are addressed by using only the ordered sequence of action labels per video during training, and forgo subjective labeling of start and end time of each action.

\begin{figure}[t]
\centering
 \includegraphics[width=0.48\textwidth,keepaspectratio]{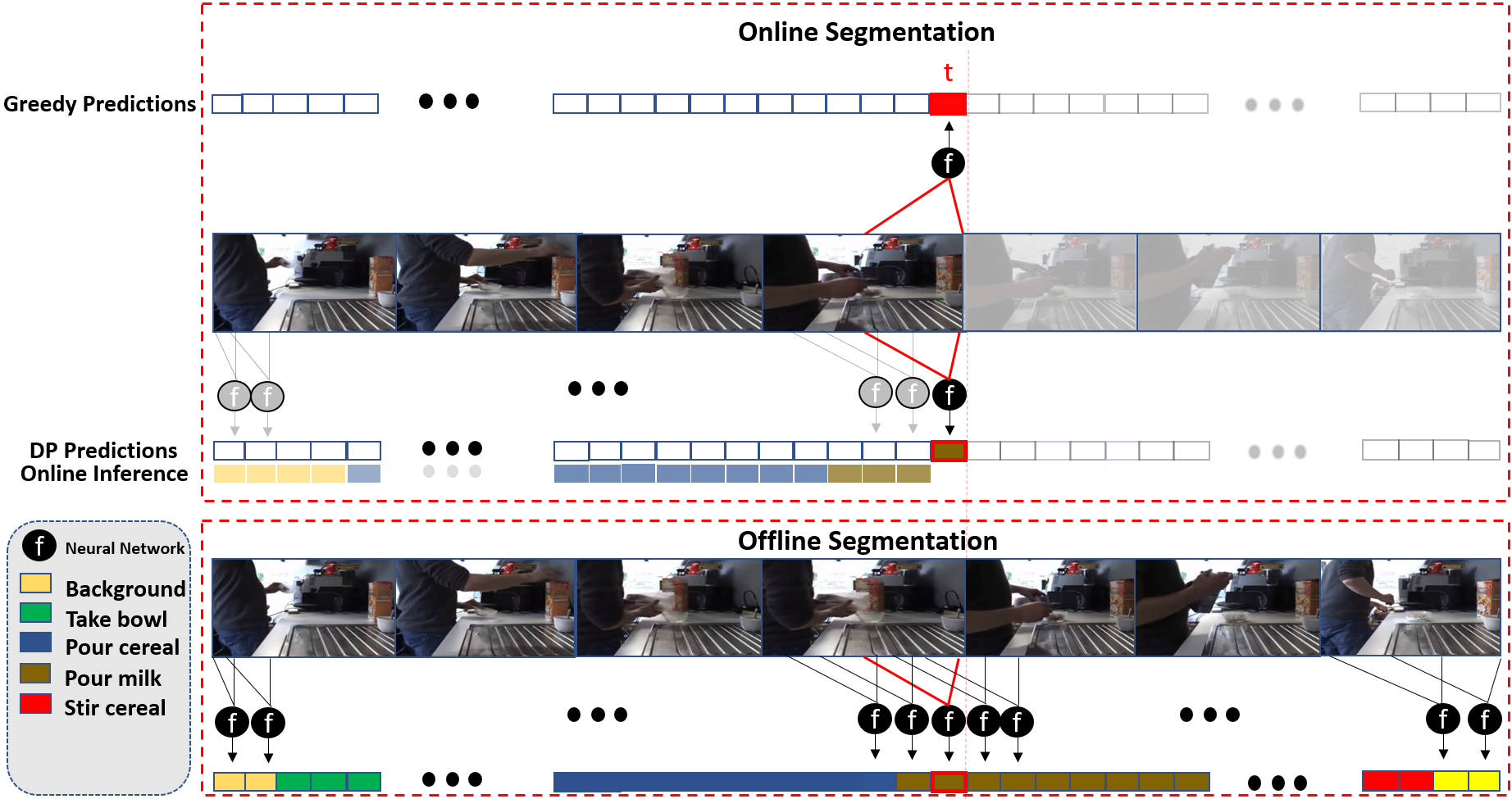}
\caption{Top: online segmentation, where the frame of interest at time t is identified either greedily by the f function or through DP-based online inference based on current and past predictions. Bottom: Offline segmentation after observing the whole sequence.}
\label{overview}
\end{figure}


Another important consideration in action understanding relates to requirements for processing the videos online versus offline, which is not addressed in existing weakly-supervised segmentation methods ~\cite{CDFL,Prototypes,MuCon}. Online processing with low latency is an increasingly important part of interactive applications where real-time, or near real-time feedback is critical.  For example, interactive applications such as human-robot interaction, error correction in manufacturing assembly, and virtual rehabilitation require immediate feedback from the intelligent system as the video streams arrive. 


The work presented in this paper considers the two aforementioned aspects in action segmentation:  weak-supervision and online processing aimed at temporally partitioning videos into action segments. To our knowledge, our work is the first to address the problem of weakly supervised online action segmentation.  Specifically, we present a framework to segment streaming instructional videos online at test time using Dynamic Programming (DP). We show the advantages of using DP as opposed to the greedy sliding window approach that are frequently used in previous online action understanding work~\cite{WOAD,eun2020learning,xu2019temporal} (Fig. \ref{overview}). 


We also introduce the Online-Offline Discrepancy Loss (OODL). Offline segmentation refers to inference after observing the video in its entirety. Offline segmentation is a non-causal procedure that is generally expected to be more accurate than its online counterpart that makes inference from partial observations. Indeed, there is a trade-off between accuracy of the recognized actions and low-latency ( Sec.\ref{latency_sec}). 
The OODL loss uses the offline segmentation result as a reference and penalizes its difference with online segmentation results generated at each time step in the video. Effectively, this encourages the segmentation results inferred at different observation end points in the video to have higher temporal consistency with respect to each other.



Furthermore, due to lack of frame-level annotation in weakly-labeled videos, frame-wise correspondence between multiple synchronized views of the same recording can provide helpful cues about the temporal location of each action during training. 
Our work is the first to use the supervision of frame-level correspondence between different views for action segmentation. We compare three ways to exploit this multi-view correspondence to generate more accurate frame-level pseudo ground-truth for weakly-labeled videos. This is in contrast to previous segmentation methods~\cite{CDFL,D3TW,Prototypes}, where different views are treated independently, discarding important multi-view information. Note that we only use the multi-view correspondence at training time and our method segments each video independently at test time with no access to other views. Also, our framework utilizes no additional annotation cost, as it is trained independent of the label and number of view points.

In summary, our main contributions are as follows:

1) We are the first to address the problem of weakly-supervised online action segmentation in instructional videos, and offer a DP-based framework.

2) We introduce the Online-Offline Discrepancy Loss (OODL). The OODL loss utilizes the offline segmentation result as a reference to train the online model by minimizing the difference between online and offline inference results.

3) We use frame-wise multi-view correspondence, during training only, to generate more accurate action pseudo-ground-truth in weakly-labeled videos with no additional annotation cost. Our work is the first to incorporate multi-view video understanding in action segmentation.

4) We present results and a detailed ablation study on two benchmark multi-view datasets in domains of cooking and assembly: Breakfast~\cite{Breakfast} and IKEA~\cite{IKEA}. We show quantitatively and qualitatively how our contributions consistently improve various suggested baselines on both datasets.

\section{Related Work} 

\textbf{Weakly-Supervised Action Segmentation.}
There has been extensive research in action segmentation of instructional videos under different forms of supervision, including fully-supervised~\cite{gao2021global2local,mstcn,fullysupervised1,wang2020boundary,TAB}, unsupervised~\cite{sarfraz2021temporally,kumar2021unsupervised,sener2018unsupervised}, and time-stamp supervised~\cite{time_stamp} methods. Methods most similar to ours use only the sequence of action labels as the weak supervision in training ~\cite{CDFL,NNViterbi,TCFPN,Fine2Coarse,D3TW,Prototypes,FIFA,DurNet}. However, all previous methods consider offline segmentation of videos, where future frames are used to make predictions at the current frame. Specifically, ~\cite{TCFPN} encodes the entire video first before decoding it to frame-level action scores.  The work in ~\cite{CDFL,NNViterbi,D3TW,Prototypes,MuCon} use Dynamic Programming (DP) to infer the most likely actions and their duration given the entire video.  Our method also uses a DP-based framework, but to our knowledge, we are the first to introduce a weakly-supervised method to segment a streaming video in an online manner.

\textbf{Online Action Understanding.}
Online action understanding has been studied in various problems such as online action detection~\cite{distill,eun2020learning,xu2019temporal}, start of action detection~\cite{gao2019startnet,shou2018online} or anticipation~\cite{gao2017red,mahmud2017joint,abu2019uncertainty,ke2019time,furnari2019would}. In the context of online action detection, ~\cite{distill} employs knowledge distillation to transfer information from offline to online models and ~\cite{qu2020lap,eun2020learning,xu2019temporal,xu2021long} introduce new neural networks to classify current actions in streaming videos using a sliding window approach. 
Others have focused solely on detecting the start of an ongoing action immediately~\cite{gao2019startnet,shou2018online} or with a short delay~\cite{buffer}. However, past methods did not consider instructional videos and, more importantly, required frame-level labels to train.

Most similar to our work is WOAD~\cite{WOAD} as the only weakly-supervised online action detection framework.
WOAD~\cite{WOAD} is different to our framework in two main ways: First, as a detection model, it is formulated to identify and localize occurrences of, typically, a single action in the input video, while we focus on instructional videos with a series of many unique actions. Second, during test time, we utilize Dynamic Programming and show in our experiments that it outperforms the greedy approach taken in ~\cite{WOAD}.

\textbf{Multi-View Action Understanding.} Using  video feeds from multiple view points has improved performance for different problems such as action recognition ~\cite{wang2018dividing,vyas2020multi,liu2020generative,piergiovanni2021recognizing,wang2019generative}, person identification ~\cite{fan2017identifying}, anomaly detection ~\cite{deepak2021deep}, and video summarization~\cite{ho2018summarizing,meng2017video,panda2017multi}. Similar to our work, ~\cite{vyas2020multi,piergiovanni2021recognizing,LAV,sigurdsson2018actor} limit exploiting multiple views to training time only. In particular, ~\cite{piergiovanni2021recognizing,vyas2020multi} focus on fully-supervised learning of trimmed videos. Meanwhile, ~\cite{LAV}  explores unsupervised video-to-video alignment, but utilize partial frame-level labels for classification. In addition, ~\cite{sermanet2018time,sigurdsson2018actor} study domain adaptation across 3rd and 1st person views. However, unlike us, they rely on view-specific labels for training.
Others ~\cite{xiong2021multiview,liu2020generative,wang2019generative} use multiple data modalities as view points. Specifically, ~\cite{xiong2021multiview} introduces a semi-supervised and view-agnostic framework for trimmed video classification, where multiple modalities are fused to generate video pseudo labels. These pseudo labels are used along with a selected number of ground-truth labels to train a video classifier.
In contrast, to our knowledge,  we are the first to use multi-view for temporal segmentation in untrimmed videos without frame-level supervision. 

\section{Background}
This section describes definitions and background concepts used henceforth. For more clarity, the supplementary material provides a table of all symbols used.
\subsection{Problem Definition}
During training, the input to our model is a video of length $T$ represented by frame-level features $\mathbf{x}_1^{T}$ and an ordered sequence of actions $\boldsymbol{\tau} =(\tau_1,\tau_2,...,\tau_{M})$ known as the transcript. $M$ is the number of actions in a given video and can vary across videos. Information about the start and end time of each action is not known.  

At test time, given the set of action labels in the dataset $\mathbb{A}$, the goal is to identify the action label $a_t\in \mathbb{A}$ at frame $t$ for all $0<t<T+1$ based on only the past and current observations $\mathbf{x}_1^t$. The final result will be a sequence of $N$ predicted segments identified online by their action $a_n$ and duration $l_n$, where $n$ refers to the $n_{th}$ segment.
\subsection{Offline Inference}
Given the input features $\mathbf{x}_1^T$ of the entire video, a common factorized formulation~\cite{NNViterbi,CDFL} to model the posterior probability of the sequence of actions $\boldsymbol{a}_1^N$ and their corresponding duration $\boldsymbol{l}_1^N$ is given by:
\begin{small}
\begin{align}
p_{\text{off}}(\boldsymbol{a}_1^N, \boldsymbol{l}_1^N | \mathbf{x}_1^{T})	\approx {}& p(\mathbf{x}_1^{T} | \boldsymbol{a}_1^N) p(\boldsymbol{l}_1^N | \boldsymbol{a}_1^N) p(\boldsymbol{a}_1^N).
\label{eq:posterior}
\end{align}
\end{small}
To infer the most likely sequence of actions $\overline{\boldsymbol{a}}_1^N$ and their duration $\overline{\boldsymbol{l}}_1^N$  associated with the video transcript $\boldsymbol{\tau}$, we use 
\begin{small}
\begin{align}
(\overline{\boldsymbol{a}}_1^N,\overline{\boldsymbol{l}}_1^N) ={}&\argmax_{\boldsymbol{a}_1^{N},\boldsymbol{l}_1^{N}}\big\{
p_{\text{off}}(\boldsymbol{a}_1^{N},\boldsymbol{l}_1^{N}| \mathbf{x}_1^{T})\big\}, \\ 
={}& \argmax_{\boldsymbol{a}_1^N,\boldsymbol{l}_1^N} \{ \displaystyle\prod_{t=1}^{T}p(x_t|a_{n(t)}) \displaystyle\prod_{n=1}^{N}p(l_n|a_n)  p(\boldsymbol{a}_1^N) \},
\label{eq:viterbi}
\end{align}
\end{small}
where $n(t)$ is the segment number at frame $t$. While training, $\overline{\boldsymbol{a}}_1^N=\boldsymbol{\tau}$ and $N=M$, since the sequence of action labels is already given in the transcript.  $p(x_t | a)$ is modeled by a GRU~\cite{GRU} and the Bayes rule as in \cite{NNViterbi}. The GRU can be optionally replaced by any other neural network as a black box. $p(l | a)$ is a Poisson distribution modeling the duration of a given action and is parameterized by the mean length of action $a$. Finally, $p(\boldsymbol{a}_1^N)=1$ if the sequence of action labels $\boldsymbol{a}_1^N$ exist in the training set transcripts and 0 otherwise. 

\subsection{Offline Segmentation Energy Score}\label{offline energy scores}
We revisit the definition of \textit{energy score} $\mathcal{E}$ introduced in offline segmentation~\cite{CDFL}. Specifically, based on the inferred segments (Eq.\ref{eq:viterbi}), we define $(\overline{\boldsymbol{a}}_1^N,\overline{\boldsymbol{l}}_1^N)$ as the unique valid path $\pi^+$ and ($\grave{a}_1^N$,$\overline{\boldsymbol{l}}_1^N)$ as an invalid path $\pi^-\in \mathbb{P}^-$, where $\grave{a}_n\in \mathbb{A} \diagdown \{\overline{a}_n\}$ and $\mathbb{P}^-$ is the set of all invalid paths given the inferred durations $\overline{\boldsymbol{l}}_1^N$. Accordingly, we define the segment-level energy score of the valid action $\overline{a}_n$ with length $\overline{l}_n$  at segment $n$ as $e_n(\overline{a}_n,\overline{l}_n)=\prod_{t \in \eta(n)}^{\eta(n)+\overline{l}_n-1} p(\overline{a}_n|x_t)$, and the segment-level energy score of an invalid action $\grave{a}_n$ is given as $e_n(\grave{a}_n,\overline{l}_n)=\prod_{t \in \eta(n)}^{\eta(n)+\overline{l}_n-1} p(\grave{a}_n|x_t)$. Here, $\eta(n)$ is a function that maps an input segment number to the starting frame number of that segment. Note that the start of each segment occurs immediately after the end of the previous segment and $p(a|x_t)$ is the output of the GRU. Further, in order to exclusively focus on hard invalid actions, the segment energy score of hard invalid actions denoted by $e_n^{-}(\grave{a}_n,\overline{l}_n)$ is defined as follows:
\[
\begin{small}
    e_n^{-}(\grave{a}_n,\overline{l}_n)= 
\begin{cases}
    \displaystyle  e_n(\grave{a}_n,\overline{l}_n),& \text{if } \displaystyle  e_n(\grave{a}_n,\overline{l}_n) > e_n(\overline{a}_n,\overline{l}_n)\\
    1,              & \text{otherwise} 
\end{cases}.
\end{small}
\]

Finally, $\mathcal{E}_{\pi^+}=\prod_{1}^{N} e_n(\overline{a}_n,\overline{l}_n)$ is the energy score of the valid path, and $\mathcal{E}_{\pi^-}=\prod_{1}^{N} e_n^{-}(\grave{a}_n,\overline{l}_n)$ is the energy score of the invalid path $\pi^-$. Calculation of these energy scores is done in the log space using DP as explained in ~\cite{CDFL}. 




\section{Weakly-Supervised Online Segmentation}
In this section, we introduce our framework for causal action inference and present how the relation between online and offline inference is exploited to derive a loss function for weakly-supervised online action segmentation.
\subsection{Online Inference}
Since online action inference is a causal process, we cannot directly use Eq. \ref{eq:viterbi} to infer the action label at the current frame ${t}^{\prime}$.
A straightforward causal solution is to employ the GRU in a sliding window fashion and apply $\argmax{\{p(a_{{t}^{\prime}}| x_t)\}}$ as the output of the GRU with the highest probability~\cite{WOAD}. However, as shown in Fig.\ref{fig:oodl_temp}, this greedy approach does not consider the context and predictions of previous time steps and is therefore sub-optimal. In order to fully account for the past actions and their duration, we formulate the marginal  causal (or online) probability $p_{\text{on}}(a_{{t}^{\prime}}| \mathbf{x}_1^{{t}^{\prime}})$ of the present action $a_{{t}^{\prime}}=a_{n({t}^{\prime})}$ at segment  $n^{\prime}=n({t}^{\prime})$ over all previous actions $\boldsymbol{a}_1^{n^{\prime}-1}$ if $n^{\prime}>1$, and duration $\boldsymbol{l}_1^{n^{\prime}}$. The inferred present action  $\hat{a}_{{t}^{\prime}}$ is derived as follows:
\begin{align}
\hat{a}_{{t}^{\prime}}={}& \argmax_{a_{{t}^{\prime}} \in \mathbb{A}} \big\{p_{\text{on}}(a_{t^{\prime}}| \mathbf{x}_1^{{t}^{\prime}})\big\}, \\
={}& \argmax_{a_{n^{\prime}} \in \mathbb{A}}  \big\{ \displaystyle \sum_{\substack{ \boldsymbol{a}_1^{n^{\prime}-1} , \boldsymbol{l}_1^{n^{\prime}}} }p_{\text{on}}(\boldsymbol{a}_1^{n^{\prime}}, \boldsymbol{l}_1^{n^{\prime}}|\mathbf{x}_1^{{t}^{\prime}}) \big\}. \label{eq:sum_rule}
\end{align}
To improve computational efficiency, we empirically approximated Eq. \ref{eq:sum_rule} by the maximum joint probability value:
\begin{align}
\hat{a}_{{t}^{\prime}}\approx \argmax_{a_{n^{\prime}} \in \mathbb{A}} \big\{ \max_{\substack{ \boldsymbol{a}_1^{n^{\prime}-1} , \boldsymbol{l}_1^{n^{\prime}}}
}p_{\text{on}}(\boldsymbol{a}_1^{n^{\prime}}, \boldsymbol{l}_1^{n^{\prime}}|\mathbf{x}_1^{{t}^{\prime}}) \big\}.
\label{eq:marginal}
\end{align}
Eq.\ref{eq:marginal} involves two steps. The first is to find the most likely sequence of actions $\tilde{\boldsymbol{a}}_1^{n^{\prime}}$ with duration $\tilde{\boldsymbol{l}}_1^{n^{\prime}}$ until time $t^{\prime}$.  The second involves taking only the last segment label $\tilde{a}_{n^{\prime}}=pop(\tilde{\boldsymbol{a}}_1^{n^{\prime}})$ to infer the label of the current frame ${t}^{\prime}$, where $pop()$ is a function that outputs the last element of a list. To execute the first step, online inference of the most likely sequence of past action segments $(\tilde{\boldsymbol{a}}_1^{n^{\prime}},\tilde{\boldsymbol{l}}_1^{n^{\prime}})$ is formulated as $\displaystyle\argmax \big\{
p_{\text{on}}(\boldsymbol{a}_1^{n^{\prime}},\boldsymbol{l}_1^{n^{\prime}}| \mathbf{x}_1^{t^{\prime}})\big\}$, where $p_{\text{on}}(\boldsymbol{a}_1^{n^{\prime}},\boldsymbol{l}_1^{n^{\prime}}| \mathbf{x}_1^{t^{\prime}})$ for $n^{\prime}>1$\footnote{For $n^{\prime}=1$, the Poisson factor $p(l|a)$ is excluded.} is derived below:
\begin{small}
\begin{multline}
p_{\text{on}}(\boldsymbol{a}_1^{n^{\prime}},\boldsymbol{l}_1^{n^{\prime}}| \mathbf{x}_1^{t^{\prime}})=\\
\Gamma(l_{n^{\prime}}|a_{n^{\prime}}) 
\displaystyle\prod_{t=1}^{t^\prime}p(x_t|a_{n(t)})  \displaystyle\prod_{n=1}^{n^{\prime}-1}p(l_n|a_n)\cdot p(\boldsymbol{a}_1^{n^{\prime}}). 
\label{eq:online_viterbi}
\end{multline}
\end{small}
$p(\boldsymbol{a}_1^{n^{\prime}})=1$ if $\boldsymbol{a}_1^{n^{\prime}}$ is a sub-sequence of any of the transcripts in the training set, and 0 otherwise, and $\Gamma(l|a)$ is a half Poisson function to model the duration $l_{n^{\prime}}$ of the current action $a_{n^{\prime}}$ at the last observed segment, given by 
\[
\begin{small}
    \Gamma(l|a)= 
\begin{cases}
1              & \text{if } l<\lambda_a \\
    \frac{\lambda_a^l \text{exp}(-\lambda_a)}{l!} & \text{otherwise } 
\end{cases},
\end{small}
\]
where $\lambda_a$ is the estimated mean length of action $a$.

Inclusion of $\Gamma()$ in the online inference of the current action is essential as it accounts for the two  following cases: First, using the full Poisson distribution of Eq.\ref{eq:viterbi} to model the duration of the current observed action leads to penalizing the current actions with a short duration, $ l_{n^{\prime}}<\lambda_{a_{n^{\prime}}}$.
However, since we do not have foresight about the duration of the current segment, any conclusion about the current segment length would be premature.
Second, $\Gamma()$ still allows us to penalize the current action if its duration is longer than expected since this can be concluded solely based on the observed segment of the action. 

At test time, the final online segmentation result in a streaming video when the current time $t^{\prime}$ changes from 1 to any given time $T$ is the sequence of frame-level actions $(\hat{a}_1,..,\hat{a}_{T})$, where each $\hat{a}_{t^{\prime}}\leftarrow \tilde{a}_{n^{\prime}}=pop(\tilde{\boldsymbol{a}}_1^{n^{\prime}})$ is inferred by Eq.\ref{eq:online_viterbi} using the Viterbi algorithm.

\subsection{Online-Offline Discrepancy Loss (OODL)}
\begin{figure}[t]
\centering
 \includegraphics[width=0.47\textwidth,keepaspectratio]{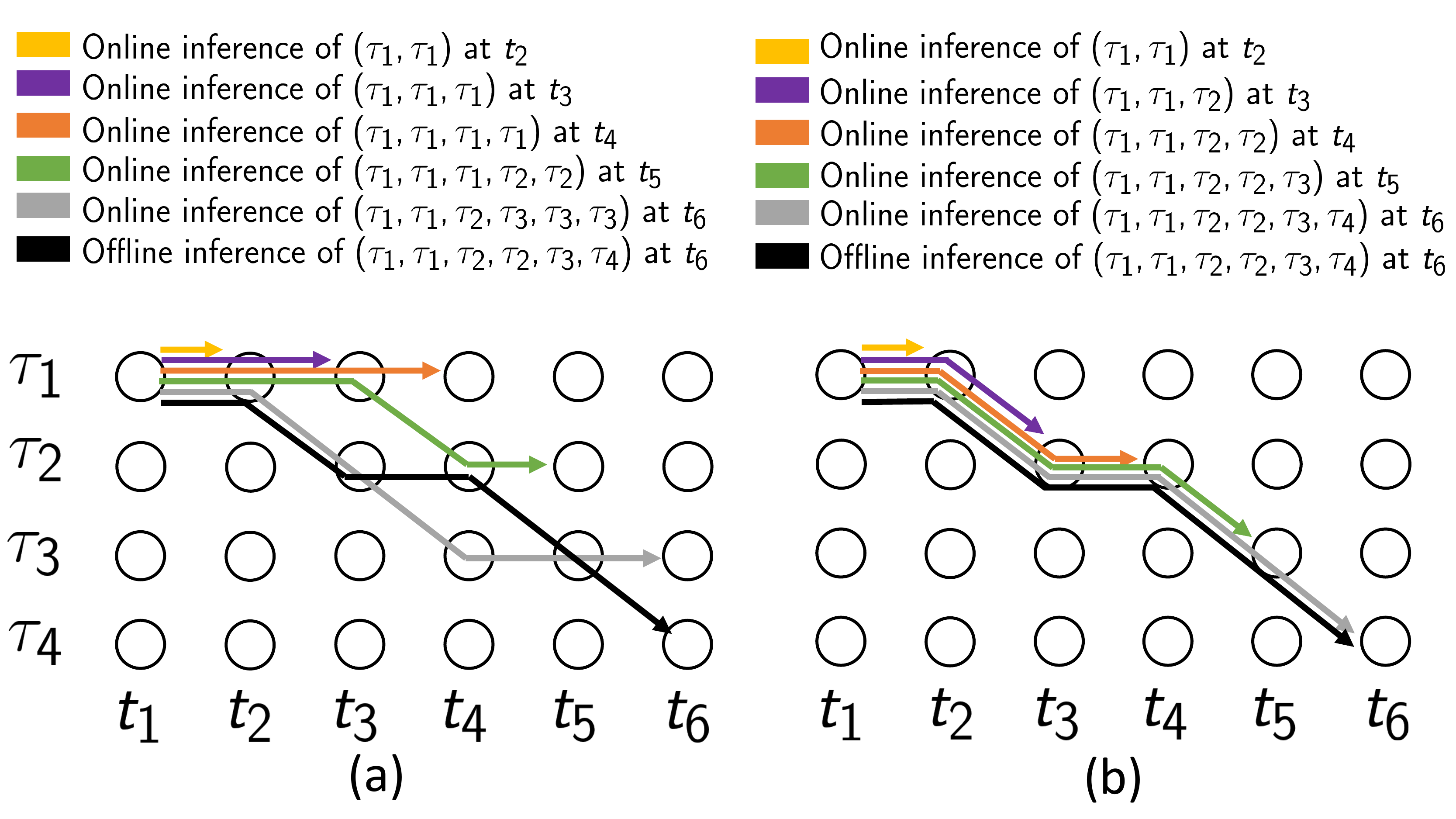}
\caption{Given the video transcript $\mathbf{\tau}$=$(\tau_1,\tau_2,\tau_3,\tau_4)$, the OODL loss encourages the online segmentation results in (a) to become a sub-sequence of the offline result for each time step as in  (b).}
\label{fig:OODL}
\end{figure}

Offline action segmentation is expected to be more accurate than online segmentation because segments are inferred from information contained in the entire length of the video, including transcripts as well as prior knowledge of the video end. Thus, offline segmentation results provide a rich source of supervision for training online segmentation models.  Ideally, the sequence of actions inferred online from the initial frame to any point in the video is expected to be a sub-sequence of the offline inference result as shown in Fig. \ref{fig:OODL}.
Consequently, this encourages all frame-level action sequences $\{\tilde{\boldsymbol{a}}_1^{t}\}_{t=1}^{T}$ to be temporally consistent, where each sequence $\tilde{\boldsymbol{a}}_1^{t}$ is inferred online at time $t$ .

We present the Online-Offline Discrepancy Loss (OODL) $\mathcal{L}_{\text{OODL}}$ in Algorithm \ref{algorithm} to minimize the difference between online and offline segmentation scores.
Specifically, we first use Eq.\ref{eq:online_viterbi} to infer the set of online paths $\{\tilde{\boldsymbol{a}}_1^{t}\}_{t=1}^{T}$ after  $n(t)$ pairs of segment-level labels $(\tilde{\boldsymbol{a}}_1^{n(t)},\tilde{\boldsymbol{l}}_1^{n(t)})$ are converted into $t$ frame-level action labels $\tilde{\boldsymbol{a}}_1^{t}$ for each time step $t$. Then, we use the hinge loss function to penalize any online inference result that has a higher energy score $\mathcal{E}_{\text{on}}$ than the energy score $\mathcal{E}_{\text{off}}$ of the offline path $\overline{\boldsymbol{a}}_1^{t} \subseteq \overline{\boldsymbol{a}}_1^{T}$  inferred from Eq. \ref{eq:viterbi}.
The OODL ultimately discourages all frame-level predictions that contribute to the discrepancy between the intermediate online inference results and the most likely sequence of actions inferred offline at the end of the video.

$\mathcal{L}_{\text{OODL}}$ is added to the baseline offline segmentation loss $\mathcal{L}_b$~\cite{CDFL} to form our final loss function $\mathcal{L}_f$:
\begin{small}
\begin{align}
\mathcal{L}_f=\mathcal{L}_b+\mathcal{L}_{\text{OODL}} .
\label{eq:final_loss}
\end{align}
\end{small}
 
Minimizing the offline segmentation loss $\mathcal{L}_b$ effectively corresponds to maximizing the decision margin between offline valid and hard invalid paths derived in Sec. ~\ref{offline energy scores}.
\begin{small}
\begin{align}
\mathcal{L}_b=-\text{log}(\mathcal{E}_{\pi^+})+ \text{log}(\displaystyle \sum_{\pi^-\in \mathbb{P}^-}\mathcal{E}_{\pi^-}).
\label{eq:CDFL_loss}
\end{align}
\end{small}

We iteratively utilize the offline and online segmentation pseudo labels inferred by Eq. \ref{eq:viterbi}, and \ref{eq:online_viterbi}, respectively, as well as the loss in Eq.\ref{eq:final_loss} to train the GRU until convergence.

\begin{algorithm}
  \caption{Online-Offline Discrepancy Loss (OODL)}
  \begin{algorithmic}[1]
    \Require{Video features $\mathbf{x}_1^{T}$ of $T$ frames and the offline inference result $  (\overline{\boldsymbol{a}}_1^{N},\overline{\boldsymbol{l}}_1^{N})$ of $N$ segments}
    \Ensure{$L_T$ as the OODL loss $\mathcal{L}_{\text{OODL}}$}
    \For{$t\gets $1 to ${T}$}:
        \State $  \tilde{\boldsymbol{a}}_1^{n(t)},\tilde{\boldsymbol{l}}_1^{n(t)}$= $ \argmax_{}\big\{p_{\text{on}}(\boldsymbol{a}_1^{n(t)},\boldsymbol{l}_1^{n(t)}| \mathbf{x}_1^{t})\big\}$ \Comment{Eq.\ref{eq:online_viterbi}}
         \State $\tilde{\boldsymbol{a}}_1^{t}$=  $(\tilde{a}_1,...,\tilde{a}_{n(t)})$ = $\text{convert}(\tilde{\boldsymbol{a}}_1^{n(t)},\tilde{\boldsymbol{l}}_1^{n(t)}) $
         \State $\mathcal{E}_{\text{on}}(t)$ = $\displaystyle\prod_{a_t\in\tilde{\boldsymbol{a}}_1^{t}}p(a_t|x_t)$
    \EndFor
         \State $\overline{\boldsymbol{a}}_1^{T}$=  $(\overline{a}_1,...,\overline{a}_{n(T)})$ = $\text{convert}(\overline{\boldsymbol{a}}_1^{N},\overline{\boldsymbol{l}}_1^{N}) $
         \State $L_0$ = $0$
    \For{$t\gets $1 to ${T}$}:
             \State $\mathcal{E}_{\text{off}}(t)$ = $\displaystyle\prod_{a_t\in\overline{\boldsymbol{a}}_1^{t}}p(a_t|x_t)$
             \State $d$ = $\max\big(0,\text{log}(\mathcal{E}_{\text{on}}(t))-\text{log}(\mathcal{E}_{\text{off}}(t))\big)$
             \State $L_t$ = $L_{t-1}$ + $\frac{d}{t}$
             \Comment{Averaging $d$ over time $t$}
    \EndFor
   \Return $L_T$
  \end{algorithmic}\label{algorithm}
\end{algorithm}

\begin{figure*}[t]
\centering
 \includegraphics[width=0.9\textwidth,keepaspectratio]{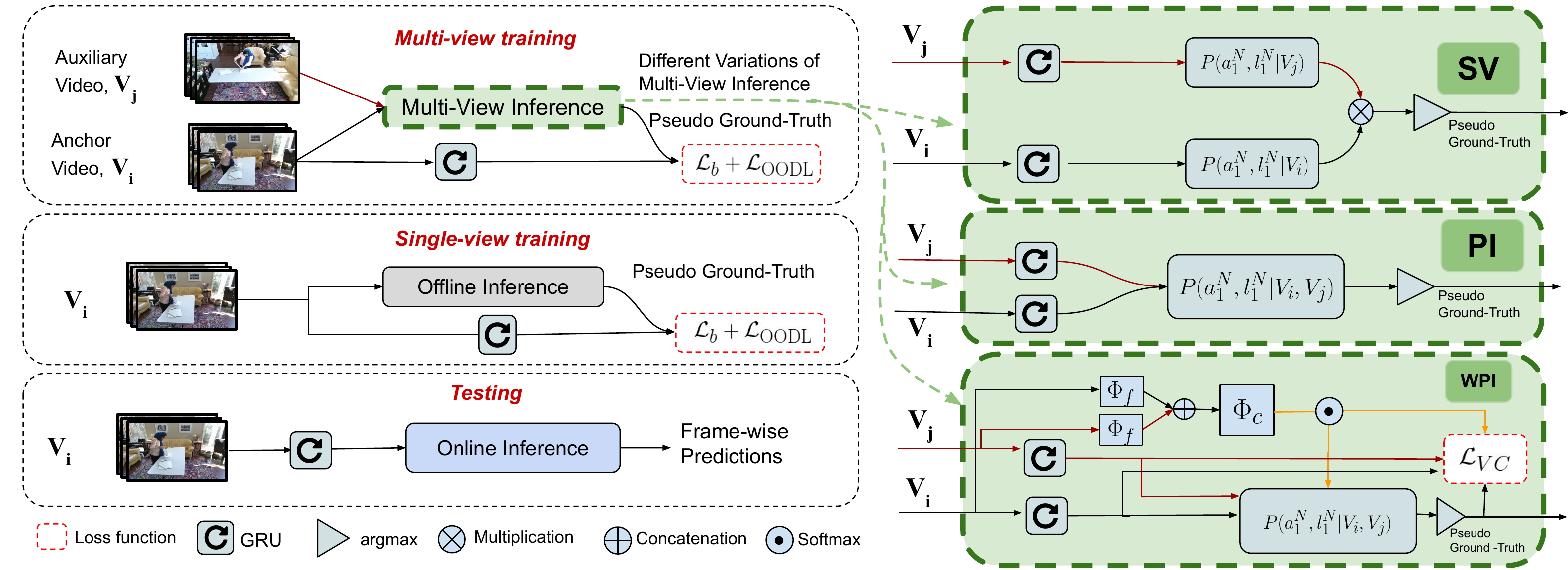}
\caption{An overview of the single-view and multi-view training schemes can be seen on the left. More details of the three proposed multi-view inference techniques are depicted  on the right. Notice how a single view is always used to segment the video at test time.}
\label{fig:multi-view}
\end{figure*}

\section{Multi-View Supervision}
Due to lack of frame-level action labels at training time, it is imperative to maximize the functional capacity of the training data available. We do so by leveraging the correspondence between multiple unknown views to infer more accurate frame pseudo labels.

Concretely, consider a training set of $K$ videos $\{v_i\}_{i=1}^K$ and their corresponding view adjacency matrix $V\in \mathbb{R}^{K\times K}$, where each element $v_{i,j}$ in $V$ is 1 if $v_i$ and $v_j$ are two different views of the same recording, and 0 otherwise. During training, we pair each video $v_i$, as the anchor video, with an auxiliary video $v_j$, which is randomly sampled from the anchor view's adjacent set $\mathbb{V}_i=\{v_k | V_{i,k}=1 \land k\neq i\}$. As shown in Fig. \ref{fig:multi-view}, each video pair is given as an input to a multiview-inference module to generate pseudo labels\footnote{Interchangeably named as offline valid path or offline inference result}, which are used to train the GRU with respect to the anchor video $i$.  In this section, we discuss three different multi-view inference techniques employed during training:

\textbf{Sequence Voting (SV).}  Given synchronized video features ${}_i \mathbf{x}_1^{T}$ and ${}_j \mathbf{x}_1^{T}$ of any two given views, we define the result of voting as the sequence of actions $\overline{\boldsymbol{a}}_1^N$ with durations $\overline{\boldsymbol{l}}_1^N$ that have the highest product of sequence probability over both views:
\begin{small}
\begin{align}
(\overline{\boldsymbol{a}}_1^N,\overline{\boldsymbol{l}}_1^N) ={}&\argmax_{\boldsymbol{a}_1^{N},\boldsymbol{l}_1^{N}}\big\{
p(\boldsymbol{a}_1^{N},\boldsymbol{l}_1^{N}| {}_i\mathbf{x}_1^{T})p(\boldsymbol{a}_1^{N},\boldsymbol{l}_1^{N}| {}_j\mathbf{x}_1^{T})\big\}.
\label{eq:SV}
\end{align}
\end{small}
In this case, the inferred sequence must have high probabilities (votes) in both views, as inconsistent probabilities (votes) diminish the overall score of any segmentation. 

\textbf{Probabilistic Inference (PI).} 
Instead of combining multi-view results at the video level as in the SV technique, here we fuse frame-level scores
to infer the sequence that maximizes the posterior  probability $p(\boldsymbol{a}_1^{N},\boldsymbol{l}_1^{N} |{}_i\mathbf{x}_1^{T},{}_j\mathbf{x}_1^{T})$ given the two views:
\begin{small}
\begin{align}
p(\boldsymbol{a}_1^N, \boldsymbol{l}_1^N | {}_i\mathbf{x}_1^{T},{}_j\mathbf{x}_1^{T}) 	\approx {}&
p({}_i\mathbf{x}_1^{T} | \boldsymbol{a}_1^N)
p({}_j\mathbf{x}_1^{T} | \boldsymbol{a}_1^N) p(\boldsymbol{l}_1^N | \boldsymbol{a}_1^N) p(\boldsymbol{a}_1^N).
\label{eq:PI}
\end{align}
\end{small}
 The $\argmax$ of the above equation can be solved by integrating  $p(x_t | a_{n(t)})=p({}_ix_t | a_{n(t)})
p({}_jx_t | a_{n(t)})$ in Eq.\ref{eq:viterbi}.

\textbf{Weighted Probabilistic Inference (WPI).} The Probabilistic Inference model in Eq.\ref{eq:PI} assumes equal contribution from each view. However, a more appropriate formulation is to compare the two views and provide a higher confidence weight on the more reliable view. Hence, we introduce  the class agnostic confidence weight ${}_ic_t\in[0,1]$ for the anchor view $i$ at time $t$ as follows:
\begin{small}
\begin{align}
{}_ic_t,1-{}_ic_t=&{}\mathrm{Softmax}\Big(\Phi_c\big([\Phi_f({}_i\mathbf{x}_{t-w}^{t}) \ \Phi_f({}_j\mathbf{x}_{t-w}^{t})]\big)\Big),
\label{eq:cv_weight}
\end{align}
\end{small}
where $\Phi_f():\mathbb{R}^{w\times F_1}\rightarrow \mathbb{R}^{F_2} $ is a function that embeds a  temporal window of the past $w$ frame features $\mathbf{x}_{t-w}^{t}$ for each view independently, and $\Phi_c():\mathbb{R}^{F_2}\rightarrow \mathbb{R}^{2}$ is the compare function that takes the concatenated view embeddings $[\Phi_f({}_i\mathbf{x}_{t-w}^{t}) \,  \Phi_f({}_j\mathbf{x}_{t-w}^{t})]$ and outputs the relative confidence weight of the anchor view $i$ with respect to the auxiliary view $j$ at time $t$. $F_1$ and $F_2$ are the dimensions of each frame feature and the window embedding respectively.

Having defined the view confidence weight ${}_ic_t$, we rewrite the likelihood $p(x_t | a_{n(t)})$ as follows and use Eq.\ref{eq:viterbi} to infer the pseudo labels $(\overline{\boldsymbol{a}}_1^N,\overline{\boldsymbol{l}}_1^N)$:
\begin{align}
p(x_t | a_{n(t)})=p({}_ix_t | a_{n(t)})^{{}_ic_t}
p({}_jx_t | a_{n(t)})^{(1-{{}_ic_t})}.
\label{eq:WPI}
\end{align}
We incorporate a new loss  $\mathcal{L}_{\text{vc}}$ in our final loss function $\mathcal{L}_{f}$ in order to learn the parameters of the view confidence weight ${}_ic_t(\boldsymbol{\theta}_c)$, where $\boldsymbol{\theta}_c$ is the set of all parameters in the compare and embedding functions of Eq. \ref{eq:cv_weight}. In addition, ${\boldsymbol\theta}_a$ denotes the set of parameters (i.e. GRU) required to predict frame-level action probability $p(a_t|x_t;\boldsymbol{\theta}_a)$.
Given the inferred pseudo labels $(\overline{\boldsymbol{a}}_1^N,\overline{\boldsymbol{l}}_1^N)$, we define the weighted energy score of the pseudo labels as :
\begin{align}
\tilde{\mathcal{E}}_{\boldsymbol{\theta}_c}
={}&\displaystyle\prod_{t=1}^{T}p(\overline{a}_{n(t)}|{}_ix_t)^{{}_ic_t({\theta}_c)}p(\overline{a}_{n(t)}|{}_jx_t)^{(1-{{}_ic_t({\theta}_c)})} , 
\label{eq:weighted_energy}
\end{align} 
where we freeze $\boldsymbol{\theta}_a$ and allow $\mathcal{L}_{\text{vc}}(\boldsymbol{\theta}_c)= -\text{log}(\tilde{\mathcal{E}}_{\boldsymbol{\theta}_c})$ to be optimized with respect to the view confidence weight ${}_ic_t(\boldsymbol{\theta}_c)$, so that the weighted energy score $\tilde{\mathcal{E}}_{\boldsymbol{\theta}_c}$ of the correct path $(\overline{\boldsymbol{a}}_1^N,\overline{\boldsymbol{l}}_1^N)$ is maximized:

\begin{align}
\mathcal{L}_{f}(\boldsymbol{\theta}_c,\boldsymbol{\theta}_a)={}& \mathcal{L}_{b}(\boldsymbol{\theta}_a)+\mathcal{L}_{\text{OODL}}(\boldsymbol{\theta}_a)+\mathcal{L}_{\text{vc}}(\boldsymbol{\theta}_c).
\label{eq:WPI_loss}
\end{align}

Note that the embedding and compare functions, $\Phi_f()$ and $\Phi_c()$,  are utilized only in training. Besides, $\mathcal{L}_{b}$  and $\mathcal{L}_{\text{OODL}}$ are computed using just the anchor video after taking the multi-view inference pseudo labels  as their valid path and offline inference result respectively. 

\section{Experiments}
\textbf{Datasets.}
\textit{The Breakfast Dataset (BD)}~\cite{Breakfast} contains approximately 1.7k  cooking videos, recorded from multiple views, ranging from a few seconds to over ten minutes long. Both the angle and number of views differ across recordings. The dataset consists of 48 action labels demonstrating 10 breakfast dishes with a mean of 6.9 action segments per video. The evaluation metrics are calculated
over four splits. \textit{The IKEA ASM Dataset (IAD)}~\cite{IKEA} has 371 recordings of assembly for four types of furniture. Each assembly is recorded from three consistent view points, providing 1113 total videos. Videos in this dataset contain a dense number of action segments (mean of $\approx$ 23) per shorter videos with a mean duration of 1.9 min.
There are 32 action classes after combining the \textit{NA} and \textit{other} classes as \textit{background}. We report results over 5 splits, where each split belongs to one of the five recording environments as suggested by~\cite{IKEA}.

\textbf{Metrics.}
Similar to previous work ~\cite{TCFPN,CDFL,Prototypes}, we use four metrics to evaluate performance: 1) \textit{acc} is the frame-level accuracy averaged over all the videos. 2) \textit{acc-bg} is the frame-level accuracy without the background frames. 3) \textit{IoU} defined as the intersection over union, which is particularly useful for imbalanced datasets such as IKEA. 4) \textit{IoD} denotes the intersection interval over the detected interval averaged across all videos. This metric tends to overrate over-segmentation results. As in~\cite{TCFPN},  both \textit{IoD}  and \textit{IoU} are calculated over non-background segments.

\textbf{Implementation.}
We extracted I3D features~\cite{I3D} for the \textit{IAD} dataset using TV-L1 optical flow ~\cite{zach2007duality} on a moving window of 16 frames. Final dimensions of the features were reduced to 400 by PCA. Meanwhile, for the \textit{BD} dataset, 
we obtained the Fisher vectors ~\cite{fisher} of iDT features ~\cite{iDT} as in ~\cite{kuehne2014language}. We implemented the  embedding function $\Phi_f()$ as temporal convolution and max pooling, while two fully connected layers were used as the compare function $\Phi_c()$. Also, we set $F_2=64$ and $\omega=21$. For a fair comparison we used the same random seed in all our experiments. The model was trained for around 70k and 6k iterations on the \textit{BD} and \textit{IAD} datasets, respectively, following the training setup of ~\cite{CDFL}.

\subsection{Comparison to the Baseline Methods}

\textbf{Baselines.} We implemented the Greedy baseline following the strategy of ~\cite{WOAD}, where a recurrent network is trained using the pseudo labels generated by an offline segmentation method. At test time, the network takes a greedy approach and identifies actions in a sliding window fashion. Also, $\text{DP}_{\text{on}}$  represents the proposed online inference (Eq.\ref{eq:online_viterbi}),  and $\text{DP}_{\text{off}}$ denotes the offline segmentation baseline of Eq.\ref{eq:viterbi}. 

\textbf{Quantitative Results.}
Table \ref{table:comparison} compares the Greedy and $\text{DP}_{\text{on}}$ methods in online segmentation. The Greedy baseline shows poor performance specially in the \textit{BD} dataset largely due to poor video quality that makes isolated predictions error-prone. However, Greedy shows decisively high \textit{IoD} values. In general, high \textit{IoD} with low  \textit{IoU} indicates over-segmentation, which leads to  overrating the result.  The presence of the $\Gamma$ function in our online modeling is important. Its omission leads to about 1\% and 3\%  drop in all metrics in the \textit{BD} and \textit{IAD} datasets, respectively. The best result is achieved by including the OODL loss and multi-view training. This leads to a 2.6\% and 3.3\%   IoU improvement of the $\text{DP}_{\text{on}}$ baseline in the \textit{BD} and \textit{IAD} datasets, respectively. Overall, improvements on the \textit{IAD} dataset are better represented by \textit{IoU} since frame accuracy is dominated by the action ``\textit{spin leg}'', which occupies nearly 45\%  of the frames.  We include offline results to show the performance gap between online and offline segmentation. The smaller gap between  $\text{DP}_{\text{off }}$ and the Greedy method in the \textit{IAD} dataset highlights the challenge of weakly-supervised learning in videos with a dense number of action segments.
\begin{table}
\begin{center}
\footnotesize\setlength{\tabcolsep}{2.5 pt}
\caption {Comparison of our multi-view supervised segmentation model with various baselines in online action segmentation. $M$ refers to multi-view training.$^*$ We report the WPI and PI multi-view results for the  \textit{BD} and  \textit{IAD} datasets respectively. \label{table:comparison} } 
\begin{tabular}{l  c | c | l l l l | l l l l}
   \multicolumn{2}{c}{Training}  &{Test}  &  \multicolumn{4}{c}{Breakfast (\%)}  &  \multicolumn{4}{c}{IKEA ASM (\%)}    \\ 
\cline{1-3}\cline{4-7}\cline{8-11}  
$M$ & $\mathcal{L}_{\text{OO}}$ & Inference &  acc & acc-bg & IoU & IoD  & acc &  acc-bg   & IoU & IoD  \\   
\hline
$\times$ & $\times$ & \scriptsize{$\text{Greedy}$}~\cite{WOAD} &   20.4 &  15.9 &   7.4 &   \textbf{58.1} &  55.6  & 56.2 & 30.9 & \textbf{53.5}\\
$\times$ & $\times$ &  \scriptsize{$\text{DP}_{\text{on}}$  w/o $\Gamma$}   &  34.3 & 31.4 &  21.4 &  45.1 &52.8 &  54.6 &  30.0 &  39.3 \\    
$\times$ & $\times$ & \scriptsize{$\text{DP}_{\text{on}}$}   & 35.1 &  32.3 &  22.4 &  46.9  & 55.3 &  57.8 &  33.3 &  44.1 \\
\checkmark & \checkmark &\scriptsize{$\text{DP}_{\text{on}}^*$}   & \textbf{36.6} &  \textbf{34.7} &  \textbf{25.0} & 49.1 & \textbf{56.9}  & \textbf{59.7} &  \textbf{36.8} &  48.0\\
\hline
\checkmark & \checkmark & \scriptsize{$\text{DP}_{\text{off}}^*$}   & 50.4 &  46.8 &  33.3 &  44.9 & 60.3  & 63.5 & 41.7 &52.0\\   
\end{tabular}
\end{center}
\end{table}
\begin{figure}[b]
\centering
 \includegraphics[width=0.49\textwidth,keepaspectratio]{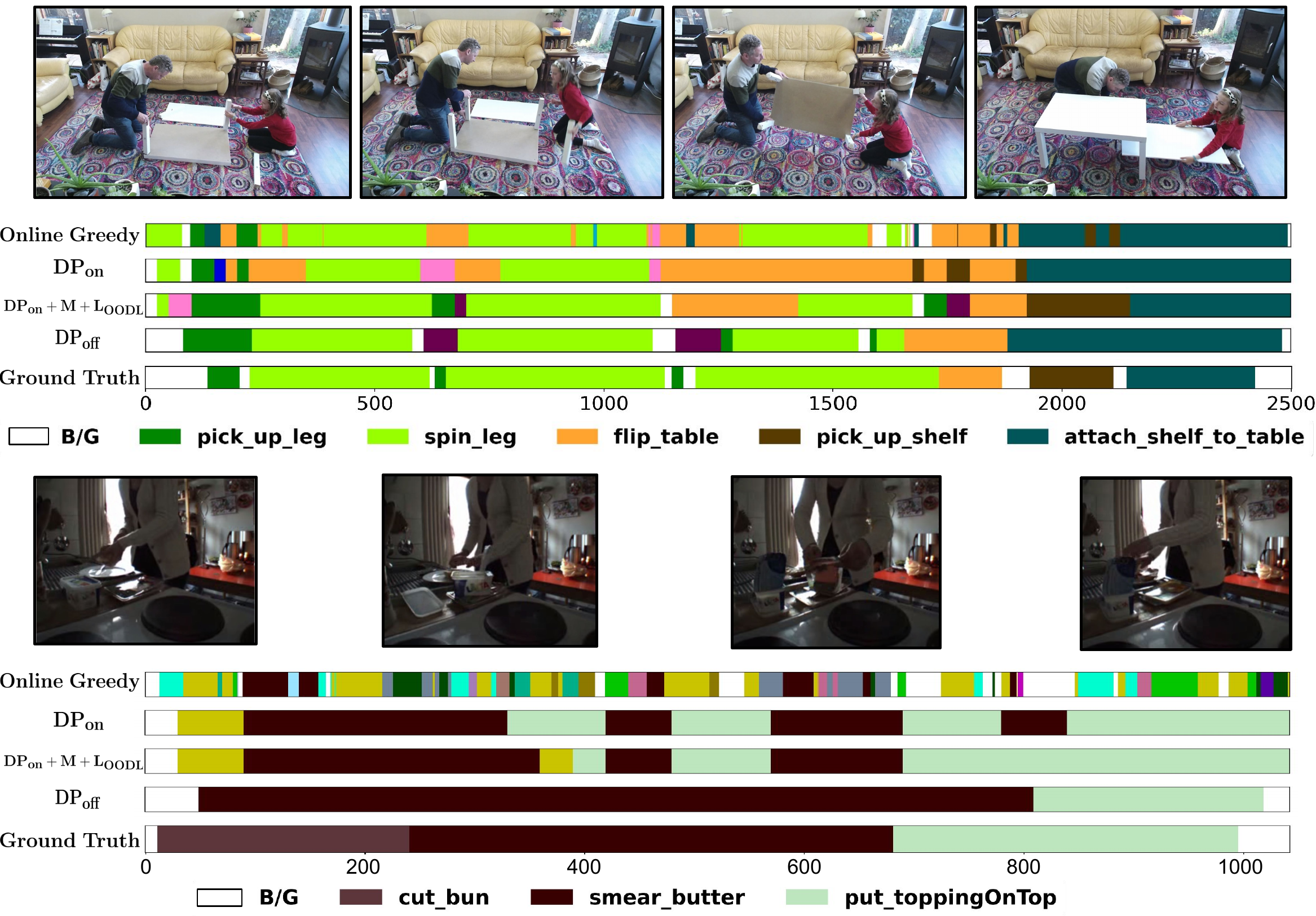}
\caption{Segmentation results of various methods on the IKEA (top) and Breakfast (bottom) datasets. Legend is shown only for the ground-truth classes.}
\label{Final_Qualitative}
\end{figure}



\textbf{Qualitative Results.} Incorporating multi-view supervision in training makes the GRU more robust to bad lighting, occlusion and scene variations as demonstrated in Fig.~\ref{Final_Qualitative}. Particularly, the top figure shows results of coffee table assembly by two people in the \textit{IAD} dataset. This is a challenging case since nearly all tasks in both datasets are completed by one person. Hence, the baseline $\text{DP}_{\text{on}}$ has missed the third instance of ``\textit{spinning the leg}'', which is correctly detected by our final segmentation model trained using $\mathcal{L}_{\text{OO}}$ and multi-view inference. The bottom figure compares different segmentation methods under dark lighting and occlusion in a sample cooking video of the \textit{BD} dataset. Notice the over-segmented results of the Greedy baseline in both cases. More examples included in the supplementary material.




\subsection{Analysis and Ablation Study}
All experiments in this section are reported as an average over all splits unless stated otherwise. Run-time and complexity, as  limitations of the proposed algorithm, are discussed in the supplementary material.
\subsubsection{Online-Offline Discrepancy Analysis}\label{latency_sec}

\begin{table}
\begin{center}
\footnotesize\setlength{\tabcolsep}{2.5 pt}
\caption {Impact of the OODL loss on weakly-supervised online segmentation results for the \textit{BD}  and \textit{IAD} datasets.\label{table:oodl} } 
\begin{tabular}{ l | l l l l | l l l l}
 &  \multicolumn{4}{c}{Breakfast (\%)}  &  \multicolumn{4}{c}{IKEA ASM (\%)}    \\ 
\cline{2-5}\cline{6-9}  
Training Approach  &  acc & acc-bg & IoU & IoD  & acc &  acc-bg   & IoU & IoD  \\   
\hline
 \scriptsize{$\text{DP}_{\text{on}}$} &  35.1  & 32.3   & 22.4   & 46.9   &  55.3   & 57.8   & 33.3   &  44.1 \\
 \scriptsize{$\text{DP}_{\text{on}}$ + $\mathcal{L}_{\text{OODL}}$} &  \textbf{35.5}  & \textbf{32.5}  &  \textbf{23.4}  &  \textbf{48.0}  &  \textbf{55.3}  & \textbf{57.9} & \textbf{34.3} & \textbf{45.5}
\end{tabular}
\end{center}
\end{table}

\textbf{Impact of the OODL Loss.}
Addition of OODL loss leads to consistent improvement in both datasets as shown in Table \ref{table:oodl}. This improvement is manifested more vividly in \textit{IoU} and \textit{IoD} because \textit{IoU}, in particular, is most appropriate in evaluating  alignment quality between predicted and ground-truth segments.

\begin{figure}[t]
\centering
 \includegraphics[width=0.47\textwidth,keepaspectratio]{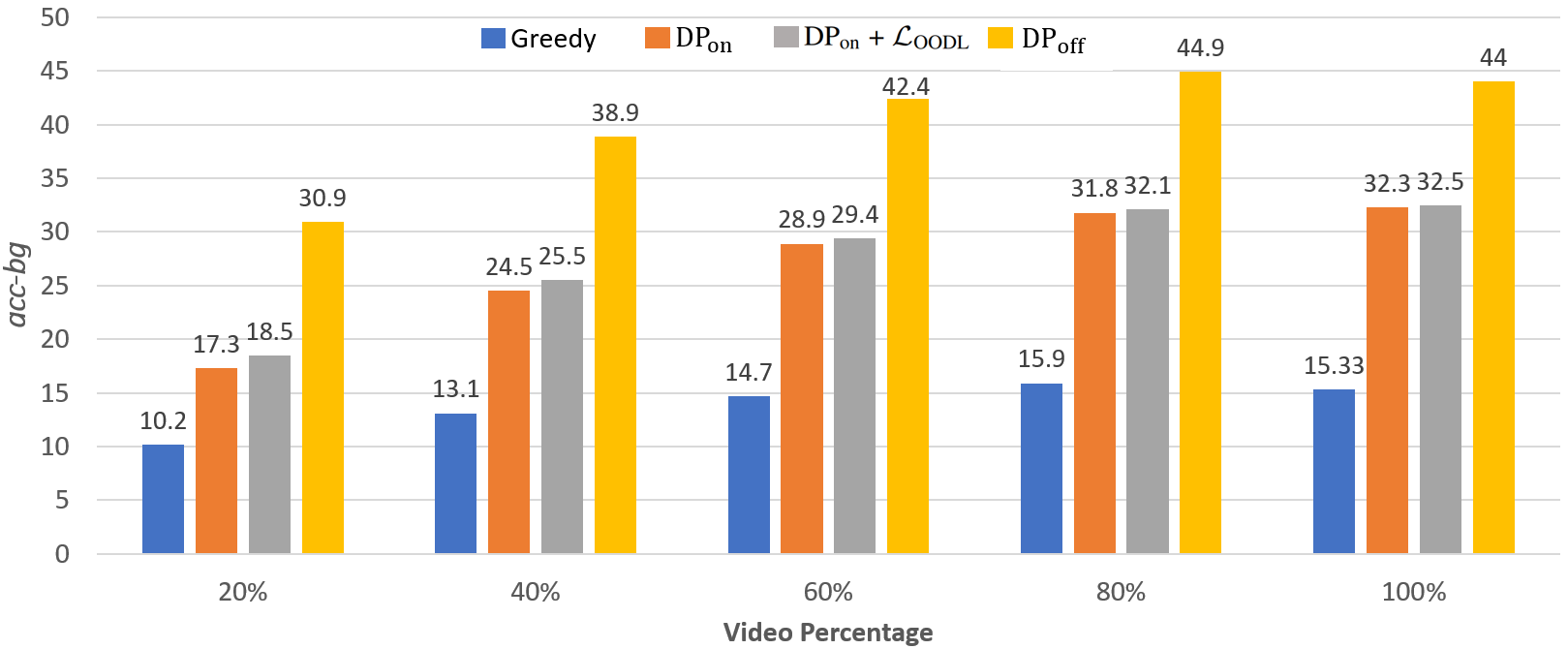}
\caption{Average segmentation results (\textit{acc-bg}) at five observation end points during the course of the video on the \textit{BD} dataset.}
\label{fig:oodl_temp}
\end{figure}

Fig. \ref{fig:oodl_temp} further demonstrates the role of the OODL loss in decreasing the online-offline segmentation discrepancy in the \textit{BD} dataset. It shows the non-background frame accuracy of multiple segmentation approaches at five different observation end points in the video. 
Upon comparison of $\text{DP}_{\text{on}}$ and $\text{DP}_{\text{on}}$ + $\mathcal{L}_{\text{OODL}}$, it can be seen that the loss has improved mostly the early predictions in the video, where it is hardest to identify actions. This is mainly due to lack of past context in early stages of a task. 
With the passage of time, more information regarding the past actions becomes available. Consequently, this leads to more accurate online predictions of the current frame. On average,  after observing the first 20\% of the video, the performance of the $\text{DP}_{\text{on}}$ is more similar to the Greedy baseline than the Offline model. However, the $\text{DP}_{\text{on}}$ approach starts resembling the Offline model more just after the 60\% point. In comparison, such a behavior is  highlighted much less by the Greedy approach due to its limitation in capturing long-range past context.


 
 

\begin{figure}[t]
\centering
 \includegraphics[width=0.48\textwidth,keepaspectratio]{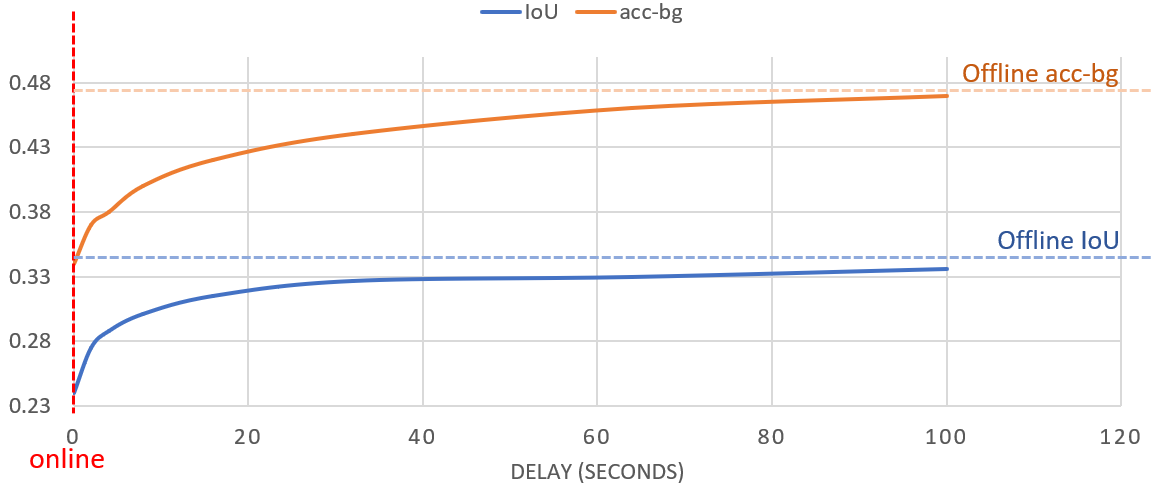}
\caption{Accuracy vs. delay on split 2 of the \textit{BD} dataset.}
\label{fig:semi_online}
\end{figure}

\textbf{Evaluation of Semi-Online Segmentation.}
Online segmentation offers practical advantages over offline inference in interactive applications that require immediate feedback. However, this comes with a 10\% and 13\% compromise over \textit{acc-bg} and \textit{IoU}, respectively, as indicated in Fig. \ref{fig:semi_online}. In some applications, certain degree of latency can be tolerated. In order to evaluate the trade-off between latency and accuracy, we implemented a semi-online variation of our framework, where predictions are made after a fixed time delay.
Fig. \ref{fig:semi_online} shows that accuracy improves with larger latency and converges to the offline result. Importantly, we can achieve approximately 90\% of the offline performance with 10 seconds of delay on the \textit{BD} dataset. 

\subsubsection{Multi-View Supervision}

\begin{table}[b]
\begin{center}
\footnotesize\setlength{\tabcolsep}{2.5 pt}
\caption {Comparison of online segmentation results under different pseudo ground-truth generation techniques (all with $\mathcal{L}_{\text{OODL}}$ ).\label{table:multi-view comparison} } 
\begin{tabular}{ l | l l l l | l l l l}
 &  \multicolumn{4}{c}{Breakfast (\%)}  &  \multicolumn{4}{c}{IKEA ASM (\%)}    \\ 
\cline{2-5}\cline{6-9}  
Training Approach  &  acc & acc-bg & IoU & IoD  & acc &  acc-bg   & IoU & IoD  \\   
\hline
 \scriptsize{Single-View} & 35.5  & 32.5  &  23.4  &  48.0   &  55.3  & 57.9 & 34.3 & 45.5 \\
  \scriptsize{SV}   & 36.4  & 34.7  &  24.8 & 48.6  & 55.7  & 58.3  & 34.6  & 45.9  \\    
 \scriptsize{PI}   & 36.2 & 34.2  & 24.4  &  48.1  & \textbf{56.9} & \textbf{59.7}  & \textbf{36.8}  & \textbf{48.0}  \\
\scriptsize{WPI}   & \textbf{36.6}  &  \textbf{34.7} & \textbf{25.0} &  \textbf{49.1} & 56.4 & 59.0 &  35.9 &  47.2\\
\hline
\scriptsize{\textit{Fully-Supervised}}   & 41.6 &  41.2 &  30.4 &  52.9 & 63.5 & 67.36 & 44.5 & 56.9 \\   

\end{tabular}
\end{center}
\end{table}

\begin{figure}[t]
\centering
 \includegraphics[width=0.46\textwidth,keepaspectratio]{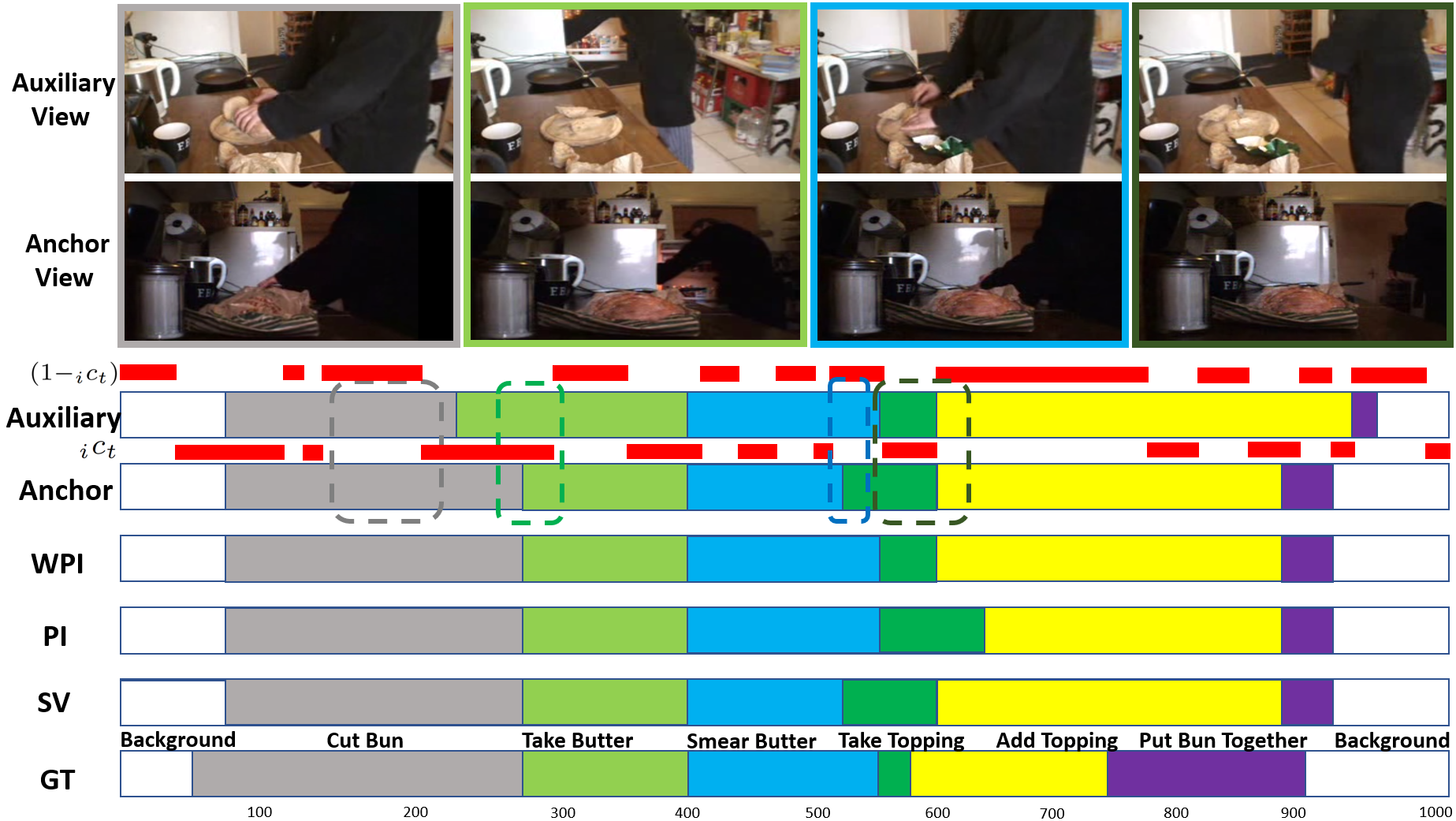}
\caption{Pseudo ground-truth generated by 3 multi-view inference techniques during training. The red bar above each single-view inference result (anchor/auxiliary) indicates their learned view confidence weight $c_t$ at each frame. Each pair of frames corresponds to the time enclosed by its color-coded dashed box.}
\label{fig:multiview_qualitative}
\end{figure}

\begin{table}
\begin{center}
\footnotesize\setlength{\tabcolsep}{2.5 pt}
\caption {Offline segmentation results  with and without multi-view supervision (same multi-view approaches as in Table\ref{table:comparison}). ~\cite{CDFL} results obtained after running the authors' code on our machine.
 \label{table:offline_segmentation} } 
\begin{tabular}{ l | l l l l | l l l l}
 &  \multicolumn{4}{c}{Breakfast (\%)}  &  \multicolumn{4}{c}{IKEA ASM (\%)}    \\ 
\cline{2-5}\cline{6-9}  
Model  &  acc & acc-bg & IoU & IoD  & acc &  acc-bg   & IoU & IoD  \\   
\hline
 \scriptsize{CDFL~\cite{CDFL}} &  49.2  & 44.2  & 31.0   & 43.7  & 59.9   & 62.0   & 39.5   &  50.4 \\
 \scriptsize{Multi-View CDFL} &  \textbf{50.4}  & \textbf{46.8}  &  \textbf{33.3}  &  \textbf{44.9}  &  \textbf{60.3}  & \textbf{63.5} & \textbf{41.7} & \textbf{52.0}
\end{tabular}
\end{center}
\end{table}

\textbf{Weakly-Supervised Online Segmentation.}
We evaluate the online segmentation performance of different multi-view inference techniques in Table \ref{table:multi-view comparison}. Regardless of the approach, using multi-view correspondence to generate pseudo ground-truth improves performance over the single-view method in all metrics and datasets. We also provide the \textit{Fully-Supervised} baseline as the upper bound, where the pseudo ground-truth is 100\% accurate.

In the \textit{BD} dataset, different views provide more complementary information as compared to the \textit{IAD} dataset. This is attributed to many instances of challenging lighting or occlusion conditions in the  \textit{BD} dataset. Fig. \ref{fig:multiview_qualitative} underscores this fact by showing the pseudo ground-truth generated during training. Specifically, the bread in the \textit{anchor} view occludes the two actions of ``cutting bread'' and ``smearing butter''. The resulting view confidence weight (red bar) of the \textit{auxiliary} view becomes high for these frames. This allows the model to exploit the more visible view in the \textit{auxiliary} video during these actions, while the \textit{anchor} view is recognized as more reliable when the subject ``takes an object''. Notice how the view confidence weights dictate the selection of single-view results to form the multi-view WPI outcome.
Meanwhile, the three views of the \textit{IAD} dataset remain fairly similar in terms of lighting and no considerable occlusion occurs in most videos. Hence, weighing views equally in the PI approach leads to the best results in the \textit{IAD} dataset.  

\textbf{Weakly-Supervised Offline Segmentation.}
Table \ref{table:offline_segmentation} shows how the advantage of multi-view training is further generalized to weakly-supervised ``offline'' segmentation. We selected the open source state-of-the-art offline segmentation method ~\cite{CDFL} as our baseline. For this experiment, we trained  ~\cite{CDFL} twice, with and without multi-view supervision, under the same parameters and random seed.


\section{Conclusion}
We introduced a framework to address a new problem of weakly supervised online action segmentation in multi-view instructional videos.  The proposed solutions are formulated with the insight that offline and multi-view results provide a rich source of supervision during training which in-turn improves performance of single view online segmentation models at test time. Extensive experiments on two benchmark datasets demonstrate efficacy of our algorithms.


\newpage

\section{Supplementary Material}

\subsection{Overview}
In this supplementary material we first provide the definitions of all the terms used in the paper, explain complexity as a limitation, and then show and discuss more qualitative segmentation results of different methods.

\subsection{Glossary of  Symbols}
We provide specific definitions of symbols in Table \ref{symbols} for readers to refer to.

\begin{table}[h]
\begin{center}
\footnotesize\setlength{\tabcolsep}{2.3pt}
\caption {Definitions of symbols used in the paper.\label{symbols} }

\begin{tabular}{| l | l |}
\hline
\textbf{Symbol} &\textbf{ Definition}  \\    \hline

\scriptsize{$\mathbb{A}$} & The set of all actions in the dataset\\
\hline
\scriptsize{$a_n$} & Action variable at segment n\\
\hline
\scriptsize{$a_t$} & Action variable at frame t\\
\hline
\scriptsize{$\hat{a}_t$} & Predicted action at time t in an online way\\
\hline
\scriptsize{$(\overline{\boldsymbol{a}}_1^N,\overline{\boldsymbol{l}}_1^N)$} & Offline inference result\\
\hline
\scriptsize{$(\tilde{\boldsymbol{a}}_1^{n(t)},\tilde{\boldsymbol{l}}_1^{n(t)})$} & Online inference result until time t\\
\hline
\scriptsize{${}_ic_t$} & View confidence weight of video i (anchor) at time t\\
\hline
\scriptsize{$e_n$} & Energy score of segment n \\
\hline
\scriptsize{$\mathcal{E}_{\pi^+}$} & Energy score of the valid path $\pi^+$\\
\hline
\scriptsize{$\mathcal{E}_{\pi^-}$} & Energy score of the invalid path $\pi^-$\\
\hline
\scriptsize{$\mathcal{E}_{\text{off}}(t)$} & Energy score of the offline or valid path until time t\\
\hline
\scriptsize{$\mathcal{E}_{\text{on}}(t)$} & Energy score of the online path until time t \\
\hline
\scriptsize{$\tilde{\mathcal{E}}$} & Weighted energy score of a path\\
\hline
\scriptsize{$F_1$} & Input feature dimension \\
\hline
\scriptsize{$F_2$} & Embedding dimension \\
\hline
\scriptsize{$K$} & Total number of videos in the data set\\
\hline
\scriptsize{$l_n$} & Duration variable of action $a_n$\\
\hline
\scriptsize{$\mathcal{L}_f$} & Final loss\\
\hline
\scriptsize{$\mathcal{L}_b$} & Baseline offline segmentation loss\\
\hline
\scriptsize{$\mathcal{L}_{vc}$} & View confidence loss to train WPI\\
\hline
\scriptsize{$M$} & Number of action labels in the transcript\\
\hline
\scriptsize{$N$} & Number of predicted segments in the video\\
\hline
\scriptsize{$p_{\text{on}}()$} & Causal probability \\
\hline
\scriptsize{$p_{\text{off}}()$} & Non-causal probability \\
\hline
\scriptsize{$\mathbb{P}^-$} & The set of all invalid paths \\
\hline
\scriptsize{$T$} & Total number of frames in the video\\
\hline
\scriptsize{$\boldsymbol{{\tau}}$} & Video transcript\\
\hline
\scriptsize{$v_i$} & Video i\\
\hline
\scriptsize{$\mathbb{V}_i$} & View adjacency set of video i\\
\hline
\scriptsize{$V$} & $K\times K$ view adjacency matrix\\
\hline
\scriptsize{$\omega$} & Feature window size for $\Phi_f$  \\
\hline
\scriptsize{$\mathbf{x}_1^{T}$} & Sequence of T frame features\\
\hline
\scriptsize{${}_i\mathbf{x}_1^{T}$} & Features of video i\\
\hline
\scriptsize{$\eta(n)$} & Mapping function from segment to frame number\\
\hline
\scriptsize{$\Gamma ()$} & Half Poisson function\\
\hline
\scriptsize{$\lambda_a$} & Estimated mean length of action $a$\\
\hline
\scriptsize{$\Phi_c$} & Compare function\\
\hline
\scriptsize{$\Phi_f$} & Feature embedding function\\
\hline
\scriptsize{$\theta_c$} & Parameters of $\Phi_f$ and $\Phi_c$ \\
\hline
\scriptsize{$\theta_a$} & Parameters of the action classifier, i.e. GRU \\
\hline
\scriptsize{$\pi^+$} & Valid path or offline segmentation action sequence \\
\hline
\scriptsize{$\pi^-$} & Invalid path \\
\hline

\end{tabular}
\end{center}
\end{table}

\subsection{Limitation}
Here we discuss the practical run-time and computational complexity of our method, both in test and training, and compare it with CDFL \cite{CDFL} as an offline baseline.  The lower frame rate compared to a greedy approach and the lengthy training-time are notable limitations of the proposed online segmentation method. We hope further work can mitigate this limitation.
\subsubsection{Runtime Frame Rate Analysis}

Both our method and the greedy approach\cite{WOAD} compute optical flow (OF) and use the I3D network to extract features. We extracted features on 320$\times$240 frames of the \textit{BD} dataset recorded at 15 fps. On a single GeForce GTX 1080, the OF and I3D network process videos at 90 and 20 fps, respectively.  Practically, if online inference is done every 15 frames, then our method segments videos at 10+ fps. While this is less than the 100 fps of the greedy alg., it leads to considerably more accurate results.

\subsubsection{Computation Complexity of Online vs. Offline}
The complexity of the proposed online inference to fully segment a video of length $T$ and maximum transcript length of $N$ is the same as the offline inference of CDFL ($O(T^2N)$). In other words,  online inference at each time step takes $O(TN)$. This is due to DP as the inference at time $t$ depends on the optimal results of previous time steps which have already been obtained as part of DP. 

With this in mind, the training complexity of our method over $K$ classes is the sum of complexities for the offline inference ($O(T^2N)$), baseline offline segmentation loss $\mathcal{L}_b$ ($O(\Delta^2NK)$) and $\mathcal{L}_{OODL}$. $\Delta \ll T$ is a small window size of 10 \cite{CDFL}.  A naive implementation of OODL has complexity of $O(T^2N)$. However, if the online and offline inferences are done together outside Alg.1 and $\mathcal{E}(t)$ is summed over segments rather than frames, the OODL complexity becomes $O(TN)$. Hence, regardless of the implementation choice, our overall training has the same complexity as CDFL ($O(T^2N+\Delta^2NK)$). Our time complexity during test time with $M$ training transcripts is $O(T^2NM)$ which is also the same as that of CDFL.  In order to quantitatively support our calculations, we tested both methods on the 4th split of the \textit{BD} dataset. Our method and CDFL took 26 and 21 hrs to train, respectively. Meanwhile at test time, ours and CDFL took 2.7 and 2.4 hrs to run, respectively.

\subsection{Qualitative Results}
\begin{figure*}[t]
\centering
 \includegraphics[width=0.9\textwidth,keepaspectratio]{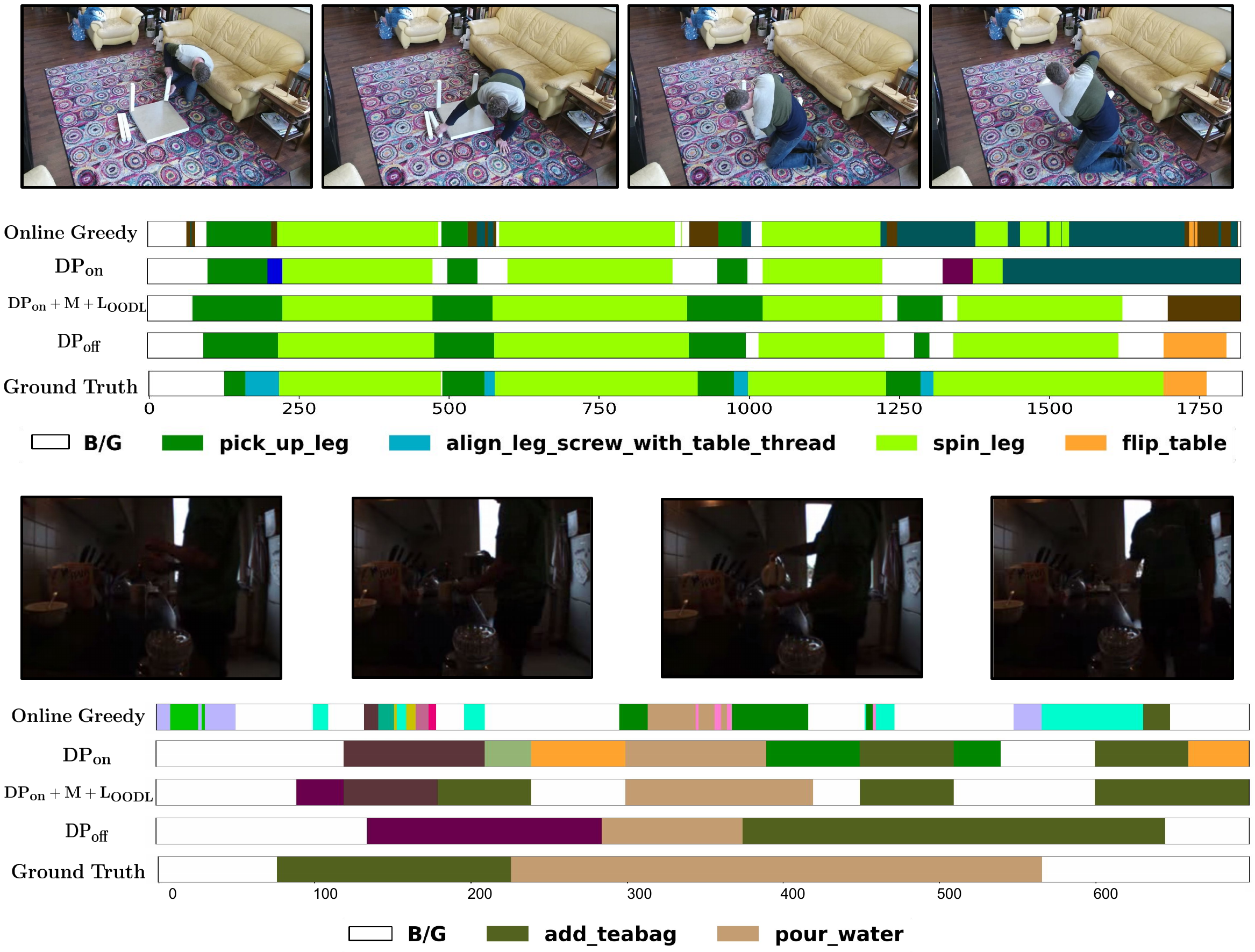}
\caption{This figure shows segmentation results of various methods on the IKEA (top) and Breakfast (bottom) datasets. Subjects in the top and bottom figures assemble a side table and prepare tea respectively. Legend is shown only for the ground-truth classes.}
\label{fig_sup:qualitative}
\end{figure*}

In Figure \ref{fig_sup:qualitative}, we present two segmentation examples on the IKEA ~\cite{IKEA} (top) and Breakfast~\cite{Breakfast} (bottom) datasets. We demonstrate how training using multiple view points has let to more robust segmentation results against full  occlusion (top) and extremely bad lighting (bottom). Specifically, the top figure depicts a task where the subject assembles a ``side table''. This assembly consists of four instances of ``spinning leg'', where the last instance is fully occluded by the subject's body. The baseline method $\text{DP}_{\text{on}}$ , that is trained on a single viewpoint,  misses most of the action, while training on multi-view correspondence and the OODL loss has enabled our final model  ($\text{DP}_{\text{on}} +M +\mathcal{L}_{\text{OODL}} $)  to capture nearly the full segment.

In the second example, the dark lighting makes it even hard for a human observer to recognize the ongoing action. Our final method is able to identify the action of ``adding tea bag'', where both the offline method $\text{DP}_{\text{off}}$ and online baselines $\text{DP}_{\text{on}}$ fail. This is an interesting case, where our model is able to outperform even the offline method. One reason is the flexibility of the proposed online segmentation model in switching between different transcripts in a series of online inferences across different time steps. This allows the predicted sequence of actions to potentially come from a transcript not observed at training time. In contrast, in offline segmentation the sequence of inferred action labels is limited only to the training transcripts.


\newpage

{\small
\bibliographystyle{ieee_fullname}
\bibliography{egbib}
}

\end{document}